# Pose Recognition in the Wild: Animal pose estimation using agglomerative clustering and contrastive learning

Samayan Bhattacharya
Department of Computer Science and Engineering
Jadavpur University
188, Raja Subodh Chandra Mallick Rd,
Jadavpur University Campus Area,
Jadavpur, Kolkata,
West Bengal 700032, India
samayan.bhattacharya@gmail.com

Sk Shahnawaz
Department of Computer Science and Engineering
Jadavpur University
188, Raja Subodh Chandra Mallick Rd,
Jadavpur University Campus Area,
Jadavpur, Kolkata,
West Bengal 700032, India
skshahnawaz2909@gmail.com

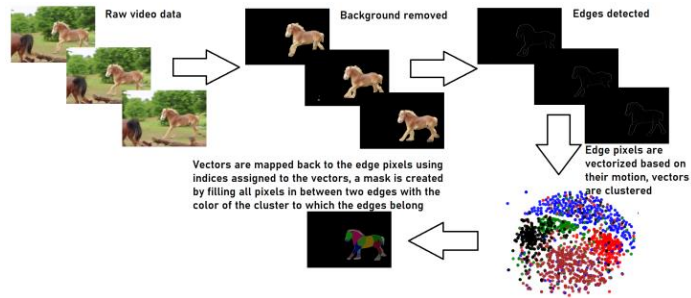

**Figure 1:** We consider the problem of unsupervised pose estimation of animals based on their motion in videos. For this, we introduce a novel method of using agglomerative clustering, with contrastive learning, on vectorized representation of edge pixels of the target animal, to allow unsupervised segmentation of the body parts of the animal. Source image credit: [51]

**Abstract**

*Animal pose estimation has recently come into the limelight due to its application in biology, zoology, and aquaculture. Deep learning methods have effectively been applied to human pose estimation. However, the major bottleneck to the application of these methods to animal pose estimation is the unavailability of sufficient quantities of labeled data. Though there are ample quantities of unlabelled data publicly available, it is economically impractical to label large quantities of data for each animal. In addition, due to the wide variety of body shapes in the animal kingdom, the transfer of knowledge across domains is ineffective. Given the fact that the human brain is able to recognize animal pose without requiring large amounts of labeled data, it is only reasonable that we exploit unsupervised learning to tackle the problem of animal pose recognition from the available, unlabelled data. In this paper, we introduce a novel architecture that is able to recognize the pose of multiple animals from unlabelled data. We do this by (1) removing background information from each image and employing an edge detection algorithm on the body of the animal, (2) Tracking motion of the edge pixels and performing agglomerative clustering to segment body parts, (3) employing contrastive learning to discourage grouping of distant body parts together. Hence we are able to distinguish between body parts of the animal, based on their visual behavior, instead of the underlying anatomy. We translate the saying, "neurons that fire together, wire together" to, "parts that move together, group together", in our approach. Thus, we are able to achieve a more effective classification of the data than their human-labeled counterparts. We run extensive ablation studies and show our results in comparison to the state-of-the-art methods. We test our model on the TigDog and WLD (WildLife Documentary) datasets, where we outperform state-of-the-art approaches by a significant margin. We also study the performance of our model on other public data to demonstrate the generalization ability of our model.*

## 1. Introduction

The focus on human pose recognition by the governments and industry has lead to the creation of humongous quantities of data. This has allowed the development of deep learning systems to segment humans [1], localize sparse 2D keypoints on humans [2], transform these 2D keypoints to 3D mesh [3],and fit complex 3D models of humans such as SMPL [4], [5],from only a picture or video of their movements. In spite of the vast importance of animal pose recognition in zoology, horticulture and aquaculture, there has been little attention paid to the non-human animals. There are 6,495 species of mammals, 60k vertebrates and 1.2M invertebrates on Earth [6]. To apply the existing methods of human pose recognition to these animals, one must bear the burden of acquiring such large, annotated datasets, as exist for human pose recognition, for each animal species. Given, the economic impracticality of such an endeavor, it is no surprise that systems for pose recognition of most species do not exist. Transferring knowledge has been proposed to adapt the annotations existing for human images to animal images [7]. This allows easy creation of annotated datasets from unannotated images and videos by mapping body parts of humans to corresponding body parts of animals. However, the prerequisite for such an operation is that, there must exist a correspondence between each body part of a human and the target animal. Thus, such methods are primarily restricted to creating datasets of primates. Another popular approach is the use of synthetic data [8]–[12]. These are easy to generate at low cost and very effective for pretraining a network. However, they are unable to reach accuracies reached with real data because they lack texture and background information, present in real images. Some works [12], [13] have explored the use of a network, trained on synthetic data to generate pseudo labels. The pseudo labels are then incorporated in training based confidence score. However, even these pseudo labels are be pretty inaccurate, leading to ineffective training. There are plenty of images and videos of animals from nature documentaries and from the work of amateur photographers, publicly available. The only reason these are not used for training networks for pose recognition tasks is that they are not annotated. Being able to make networks train on these unannotated images and videos, for pose recognition, would open up a wide range of opportunities for scientific studies of these animals. In this paper we propose a network that is able to segment body parts of an animal, based on its movements, from unannotated videos. Our method works for all animal species, including fish. To the best of our knowledge, none of the previous studies have considered pose recognition for fish. Our proposed method consists of several steps. Firstly, we remove the background information from each frame of the video. This helps the network focus on the animal alone and saves it from having to figure out which part of each frame is the background. The result of this step is the animal with black (RGB:0,0,0) background. Then we use an edge detection algorithm to isolate the outline of the body of the animal. This facilitates easy motion tracking of the animal, based on the Euclidean distance of each edge between subsequent frames of the video. The only limitation to this approach is, its performance deteriorate with increase in the ratio of (speed of movement of the animal) to (frame rate of the camera used). Second, we track the motion of the edges, based on the assumption that the edge the present frame, closest to the edge in the previous frame, is the next position of that frame. This is much simpler than other motion tracking methods in use [14], [15]. The result of this step is some vectors representing the displacements of the edge pixels. These vectors are passed to an agglomerative clustering network to be divided into clusters, each of which represents a body part. The disadvantage of this approach is that the model tends to group disconnected parts of the animal into one cluster because they have similar motions. To overcome this we use the third step. In the third step, we use a contrastive learning [16] to discourage the network from grouping distant body parts into the same cluster. To the best of our knowledge, contrastive learning have not previously been studied with agglomerative clustering. For this step we use two identical agglomerative learning networks and train them using a contrastive loss function. Thus, the network is able group disconnected body parts that move similarly into separate clusters

## 2. Related work

### 2.1. Human pose recognition

Abundant attention and funding by governments and industry has resulted in a lot of work in the domain of human pose recognition, both in 2D and 3D. Works in 3D mostly rely on keypoint extraction from 2D images, which are then mapped to 3D [17]. Deep learning has been widely used for this purpose [2], [17], [18]. These are trained on large, manually annotasted datasets, such as Posetrack [19], PennAction [20],Leeds Sports Pose Dataset (LSP) [21], [22], MPII [23], and COCO [24]. Some attempts to learn pose recognition in an unsupervised manner exist [7], [25]–[30], however, they produce poor results in comparison to the supervised techniques..

### 2.2. Animal pose recognition

There have been several works on applications of deeplearning for detection and segmentation of animals. Since images of animals are so widely spread, they appear in multiple 2D recognition datasets like the COCO dataset. However, for animal pose recognition tasks, each of the existing works focuses on one animal species or a small group of visually similar species, such as drosophila melanogaster flies[31], cheetahs [32] or Amur tigers [33]. These produce good results only for a particular animal species and a particular environment.

Several annotation tools have also been developed to map 2D keypoints to 3D, such as such as Anipose [34] and DeepLabCut [35]. However, these require specialized data, to allow multiple views and triangulation. One of the few works for animal recognition by using only visual data has been the recognition of facial landmarks by domain adaptation[36], [37]. Another work [13] focused on the full body recognition of four-legged animals by using cross-domain adaptation network, trained on a large number of annotated human images and a smaller number of annotated animal images. Another set of works [8]–[10], based on Skinned Multi-Animal Linear (SMAL), obtained from 3D scans of toy animals and having the capacity to represent multiple classes of mammal and parametric linear model, is focusing on model-based estimation of 3D pose and shape. [12] uses synthetic data generated using CAD. These are then used to generate pseudo labels for unlabelled animal images.

## 2.3. Unsupervised and less supervised pose recognition

Recent works [10], [28], [30], [31], [39] have considered learning sparse and dense visual landmarks for simple classes. These methods are unstable for most applications. [40] at-tempted to reduce the number of annotated images required for pose estimation of humans. Another approach [41]–[44] uses adversarial networks for domain adaptation. The generator learns domain invariant feature to fool the discriminator. This facilitates the transfer of labels from source to target domain

## 3. Proposed method

The goal of our work is to be able to perform segmentation of the body parts of different species of animals, based on their movement, in an unsupervised manner. Segmentation of images would allow easy annotation of the images, by simply hard coding the mapping of each segment to a label. The TigDog dataset [45] provides short videos of dogs and tigers. We use these videos without the annotations.

In this work we achieve the following tasks:
(i) perform agglomerative clustering on vector representation of the motion of edges of the animal in the video
(ii) avoid grouping of distant segments of the body using contrastive loss

## 3.1. Background removal

The background information is irrelevant to the task of animal pose detection and is a potential source of data leakage. We base the study of our proposed method on the animal alone, without any suggestion from the background. We use Robust High-Resolution Video Matting with Temporal Guidance, a recent work by [46] to remove the background. This lightweight method allows effective background removal, without additional computational burden. Once the background is removed, the video consists of the animal moving against a black background (RGB:0,0,0).

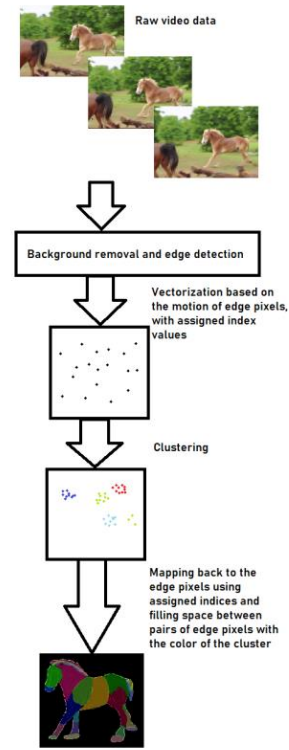

**Figure 2:** A single agglomerative clustering pipeline

## 3.2. Edge detection

Since we are focusing on the motion of the animal, the identity of the animal species is irrelevant to this task. Hence we are able to reduce computation by removing the body of the animal, while preserving only the edges. According to our experiments, the edges provide sufficient information for the purpose of motion tracking. For this we use a modified version of the Canny edge detection algorithm [47].

### 3.3. Conversion to vectors

To track the motion of the animal we employ a pixel-wise coattention mechanism. Since we have removed all parts of the image, except the edge pixels of the animal being tracked, we are able to adopt the simple approach of tracking the position of each pixel in subsequent frames of the video.

Say, the position of a particular edge pixel in one frame is (x1,y1) and its position in the next frame is (x2,y2). The Euclidean distance between (x1,y1) and (x2,y2) is assumed to be minimum among the Euclidean distances of (x1,y1) and any pixel. The accuracy of this assumption depends on the speed of movement of the animal relative to the frame rate of the camera being used. Nevertheless, the assumption holds true for the TigDog dataset and other datasets we tested in this study.

$$Euclidean\ distance = \sqrt{(x2-x1)^2 + (y2-y1)^2} \quad (1)$$

The movement of the pixel is converted to a vector, represented by [[x1,y1],[x2,y2]], for two consecutive frames. The vectors from all pairs of two consecutive frames are concatenated to form a higher dimensional vector. Thus, each edge pixel is represented by a vector that contains information about the motion of the pixel throughout the video. We assign an index number to the vector and the pixel that it represents, to allow identification.

### 3.4. Agglomerative clustering

The vector representation of the edge pixels is subjected to agglomerative clustering, based on complete linkage algorithm. This divides the vectors into separate clusters, based on their proximity to the centroid of a particular cluster. The goal of this architecture is to increase the inter-cluster distance and reduce the intra-cluster distance.

$$cluster_{centroid} = \left(\frac{\sum_{i=1}^{i=n}(x_i)}{n}, \frac{\sum_{i=1}^{i=n}(y_i)}{n} \ldots \right) \#(2)$$

and

$$intra_{distance} = \frac{\sum_{i=1}^{i=n}\sqrt{(x_i - cluster_{centroidx})^2 + (y_i - cluster_{centroidy})^2} \ldots}{n} \#(3)$$

Complete linkage algorithm is used because it handles outliers and noise points better. This step results in the edge pixels being divided into clusters, based on their common movements in the video. Since, all parts of the body of the animal share some common motions, the agglomerative clustering network is later trained with a contrastive loss, to discourage distant body parts from being grouped together.

### 3.5. Contrastive training

Contrastive loss is used in both semi-supervised [48] and unsupervised learning [49] settings. The goal of using contrastive loss is to discourage the network from producing similar outputs for dissimilar inputs and vice versa.

$$loss_{contrastive}\left(w, (y, \overrightarrow{x1}, \overrightarrow{x2})\right) = \sum_{i=1}^{i=n}((1-y).L_S.D_W + (y).L_D.D_W)\#(4)$$

Where, $\overrightarrow{x1}$, $\overrightarrow{x2}$ are two data points, y is either 0 (for similar data points) or 1 (for dissimilar data points) and $D_w$ is the difference between the data points. $L_S$ is the loss function for similar data points and $L_D$ is the loss function for dissimilar data points.

$$D_W = ||G_W(\overrightarrow{x1}) - G_W(\overrightarrow{x2})||\#(5)$$

For this study, we use two identical agglomerative clustering models, independently trained on the data. Then we train them using the contrastive loss function.

For similar data points, we sample from neighboring regions of the image, while for dissimilar examples, we sample from distant regions of the image. The mapping from the vector representation to edge pixels, using the index numbers assigned, becomes useful for this purpose.

This step allows the network to reconfigure the existing clusters as well as form new clusters, to be able to distinguish between body parts that are distant from one another. Thus, the two front legs of a running horse are put into separate clusters, even though they go through almost the same motion [50]. The performance of the model on the TigDog [45] and WLD [51] datasets also improves upon using the contrastive loss function.

### 4. Training

The agglomerative model is initially trained on the TigDog dataset. After initial training of two identical models, the two models are trained with contrastive loss. A sampling function is used for providing inputs to the models while training. The sampling function samples from neighboring regions of the image to provide similar inputs (Y=0) and it samples from distant parts of the image to provide dissimilar inputs (Y=1),. The minimum distance between the two dissimilar inputs depends on the average size of the animal body in the images and needs to be fine-tuned for a dataset.

Two agglomerative clustering models are trained independently and in a contrastive setting, alternatively.

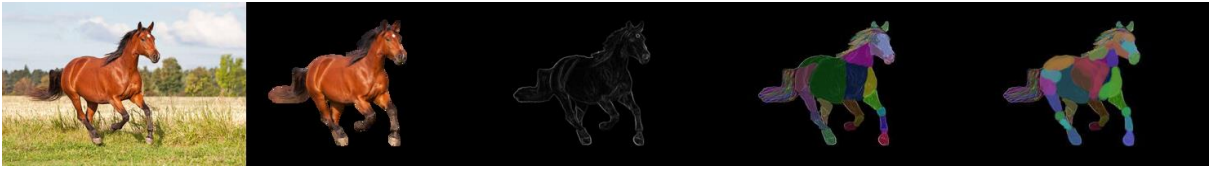

**Figure 3:** Visual results. From left to right, the images are, the original image, image after background removal, image after edge detection and the final result from clustering the edge pixels and filling all pixels in between two edge pixels with the color of the cluster to which the two edge pixels belong, the left on is with contrastive learning and the right one is without contrastive learning. Image source: [45], [51], data generated at the lab, by the authors.

This allows the network to learn the most appropriate grouping of the body parts of an animal, based on their movements.

## 5. Experiments

In this section, we provide the results of experimental evaluation and detailed description of the ablation study.

### 5.1. Datasets

We use several different datasets, consisting of videos of animals of different species. The datasets are described below.

**TigDog dataset**[45]. This dataset consists of short videos of 3 animal species: tigers, sourced from nature documentaries, horses and dogs, sourced from the YouTube-Objects dataset. Thus, the dataset is divided into 3 classes, each corresponding to an animal species. Each video contains at least one instance of the class to which it belongs. The videos are annotated according to the kind of motion the animal is performing, eg: walk, run or turn head. If a video contains different motions from different animals, only the animal closest to the camera is considered. We ignore these annotations as well as other annotations of body parts in this study.

**WLD (WildLife Documentary)** [51]. This dataset consists of videos of several species including 'tiger', 'koala', 'langur', and 'ostrich'. These are sourced from 15 documentary films on YouTube. The videos range from 9 minutes to 50 minutes in duration and 360p to 1080p in resolution. The videos have subtitles, which we disregard in this study. Each animal is annotated, resulting in about 4098 object short videos consisting of 60 different animal species.

**Random videos from YouTube**. We test our method on random videos downloaded from YouTube in order to ensure generalizabilty. Our model also performs well on videos of insects, arachnids and reptiles. Since no large dataset is available for such, highly overlooked classes of animals, we had to rely on publicly available nature documentaries as a source of such videos.

### 5.2. Results

The lack of ground truth data prevented us from finding accuracy of our model on all datasets we tested in our study. The performance of the model was thus assessed by visual inspection and mean log-likelihood values. For TigDog dataset, we obtained ground truth labels by Amazon Mechanical Turk. Hence, we are able to report Average Precision values for this dataset. We demonstrate that the model with the contrastive learning step performs better than the one without it. It is also observed that longer duration videos produce better results that shorter duration videos, for a given animal class.

To overlay the learned clusters on the original image, we assign the color of the cluster, to which two edges of a body part are assigned, to all pixels between those two edges. This mask is then overlaid on the original image to produce the final result.

Table 1: AP of the network trained on videos of length 1 min, 5mins and 10 mins with contrastive learning step. Mean±std for 20 runs.

| Dataset | $AP_1$ | $AP_5$ | $AP_{10}$ |
|---|---|---|---|
| TigDog (Tiger class) | 32.5±0.4 | 35.4±0.3 | 37.2±0.4 |
| TigDog (Dog class) | 30.2±0.5 | 33.6±0.4 | 34.6±0.3 |

Table 2: AP of the network trained on videos of length 1 min, 5mins and 10 mins without contrastive learning step. Mean±std for 20 runs.

| Dataset | $AP_1$ | $AP_5$ | $AP_{10}$ |
|---|---|---|---|
| TigDog (Tiger class) | 30.1±0.3 | 32.6±0.3 | 33.4±0.6 |
| TigDog (Dog class) | 27.2±0.5 | 30.1±0.2 | 31.7±0.4 |

Table 3:Mean log-likelihood estimates, for the predicted segmentations, when trained on videos of length 1 min,

5mins and 10 mins with contrastive learning step. Mean±std for 20 runs.

| Dataset | $AP_1$ | $AP_5$ | $AP_{10}$ |
|---|---|---|---|
| TigDog (Tiger class) | 220±1.6 | 240±2 | 250±0.5 |
| TigDog (Dog class) | 200±2.1 | 210±1 | 230±1.5 |
| WLD (Tiger class) | 215±1.3 | 240±1.2 | 246±2 |
| WLD (Koala class) | 230±1.5 | 250±2 | 255±1.9 |
| WLD (Langur class) | 200±2 | 210±1.4 | 230±0.6 |
| WLD (ostrich class) | 220±2 | 225±1.6 | 230±2 |
| Nature documentary (Spider class) | 190±0.4 | 199±0.9 | 210±0.9 |
| Nature documentary (Beetle class) | 165±0.7 | 174±1.2 | 190±1.6 |
| Nature documentary (Snake class) | 100±2 | 116±2.1 | 120±1.6 |

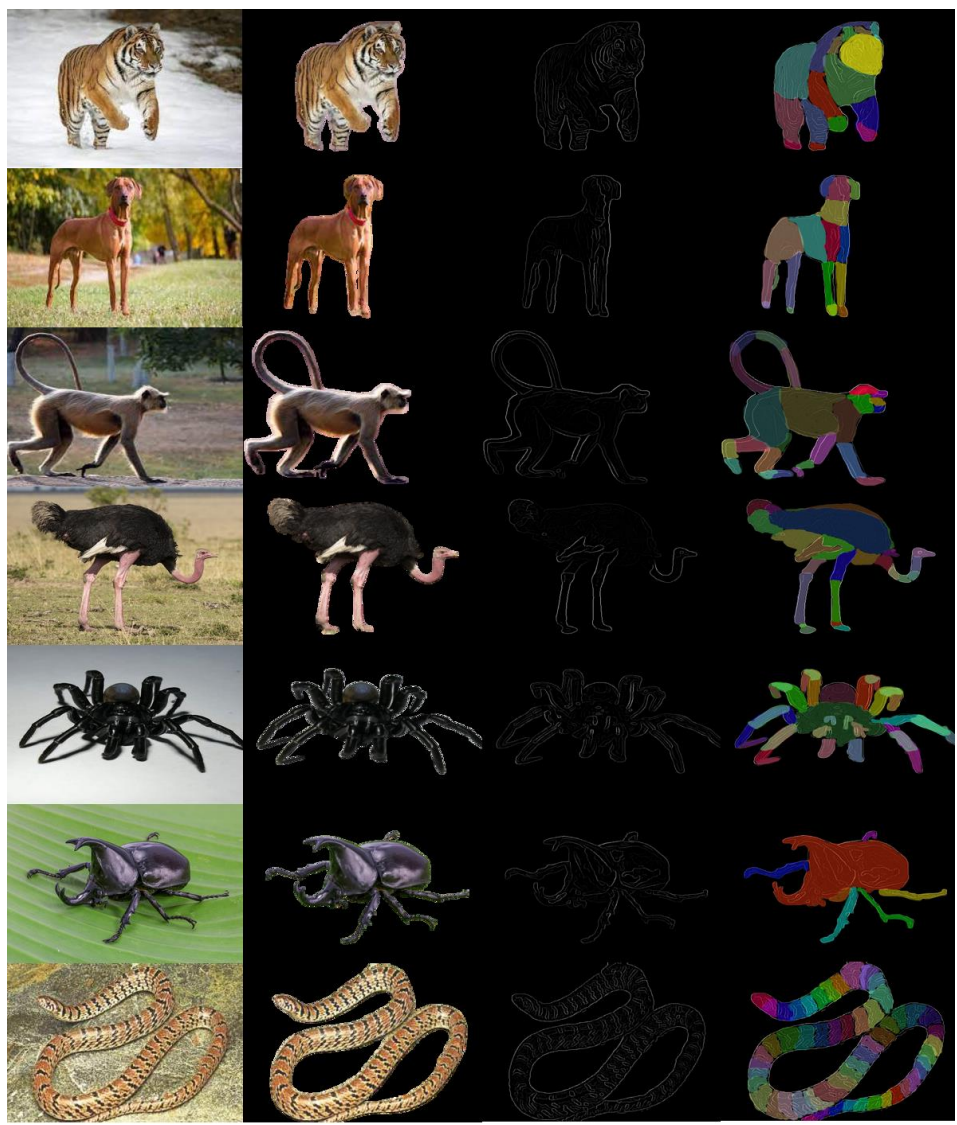

**Figure 4:** Visual results. From left to right, the images are, the original image, image after background removal, image after edge detection and the final result from clustering the edge pixels and filling all pixels in between two edge pixels with the color of the cluster to which the two edge pixels belong. This mask is overlaid on the image of the edges for clarity. Please zoom in for details. Image source: [45], [51], data generated at the lab, by the authors.

## 6. Conclusion

In this paper, we introduced a novel approach for segmenting the body of animals, based on their movements alone. We are able to achieve good mean log-likelihood values for the datasets lacking ground truth labels and good Average Precision for the TigDog dataset. The most important contribution of this work is the capability to segment the body of animals with complex body shapes, e.g.: reptiles, insects, arachnids, etc., which are largely ignored by the existing research.

As a next step, we would like to explore annotating the clusters, so as to assign a meaningful label to each segment. This would also allow mapping of these body parts to similar parts in humans or other animals with closely related anatomical features.